\documentclass[conference]{IEEEtran}

  	\usepackage[pdftex]{graphicx}
  	\graphicspath{{../pdf/}{../jpeg/}}
	\DeclareGraphicsExtensions{.pdf,.jpeg,.png}

	\usepackage[cmex10]{amsmath}
        \usepackage{amssymb}
	\usepackage{mathabx}
        \usepackage{mathrsfs}
        \usepackage{subfig}

        \usepackage{algorithm}
        \usepackage{algorithmic}
        \usepackage{authblk}
        \usepackage{hyperref}
        \usepackage{booktabs}
        \usepackage{multirow}
        \usepackage[T1]{fontenc}
        \usepackage[utf8]{inputenc}
	\usepackage{array}
	\usepackage{mdwmath}
	\usepackage{mdwtab}
	\usepackage{eqparbox}
	\usepackage{url}
        \usepackage {color}
 
\begin{document}

    \title{\LARGE Multi-Vehicle Trajectory Planning at V2I-enabled Intersections based on Correlated Equilibrium}
    \author[a]{Wenyuan Wang}
    \author[b,c,d]{Peng Yi}
    \author[b,c,d]{Yiguang Hong}
    \affil[a]{Department of Control Science and Engineering, Tongji University}
    \affil[b]{Shanghai Research Institute for Intelligent Autonomous  Systems, Tongji University}
    \affil[c]{National Key Laboratory of Autonomous Intelligent Unmanned Systems}
    \affil[d]{Frontiers Science Center for Intelligent Autonomous Systems, Ministry of Education}

    \renewcommand*{\Affilfont}{\small\it} 
    \renewcommand\Authands{ and } 
    \date{} 


    \maketitle

\begin{abstract}
Generating trajectories that ensure both vehicle safety and improve traffic efficiency remains a challenging task at intersections. Many existing works utilize Nash equilibrium (NE) for the trajectory planning at intersections. However, NE-based planning can hardly guarantee that all vehicles are in the same equilibrium, leading to a risk of collision. In this work, we propose a framework for trajectory planning based on Correlated Equilibrium (CE) when V2I communication is also enabled. The recommendation with CE allows all vehicles to reach a safe and consensual equilibrium and meanwhile keeps the rationality as NE-based methods that no vehicle has the incentive to deviate. The Intersection Manager (IM) first collects the trajectory library and the personal preference probabilities over the library from each vehicle in a low-resolution spatial-temporal grid map. Then, the IM  optimizes the recommendation probability distribution for each vehicle's trajectory by minimizing overall collision probability under the CE constraint. Finally, each vehicle samples a trajectory of the low-resolution map to construct a safety corridor and derive a smooth trajectory with a local refinement optimization. We conduct comparative experiments at a crossroad intersection involving two and four vehicles, validating the effectiveness of our method in balancing vehicle safety and traffic efficiency.
\end{abstract}


\let\thefootnote\relax\footnotetext{The paper was sponsored by the National Key Research and Development Program of China under No 2022YFA1004701,  the National Natural Science Foundation of China under Grant No. 72271187 and No. 62373283, No. 62088101 and partially by Shanghai Municipal Science and Technology Major Project No. 2021SHZDZX0100}

\section{Introduction}
Traffic intersection is a critical and challenging scenario for autonomous driving since it is an area of high-risk accidents. Franke et al. summarize that more than 33\% of injury-causing traffic accidents occur at urban intersections \cite{franke2008dynamic}. Particularly, intersections account for over 40\% of reported traffic accidents in the United States and Europe \cite{azimi2014stip}. Intersections are also a major cause of traffic congestion \cite{rafter2017traffic}, and meanwhile, accident probability at intersections increases with the level of congestion \cite{retallack2019current}. The economic and social losses at traffic intersections are substantial \cite{namazi2019intelligent}.

\begin{figure}[ht!] 
\centering
\includegraphics[width=3.5in]{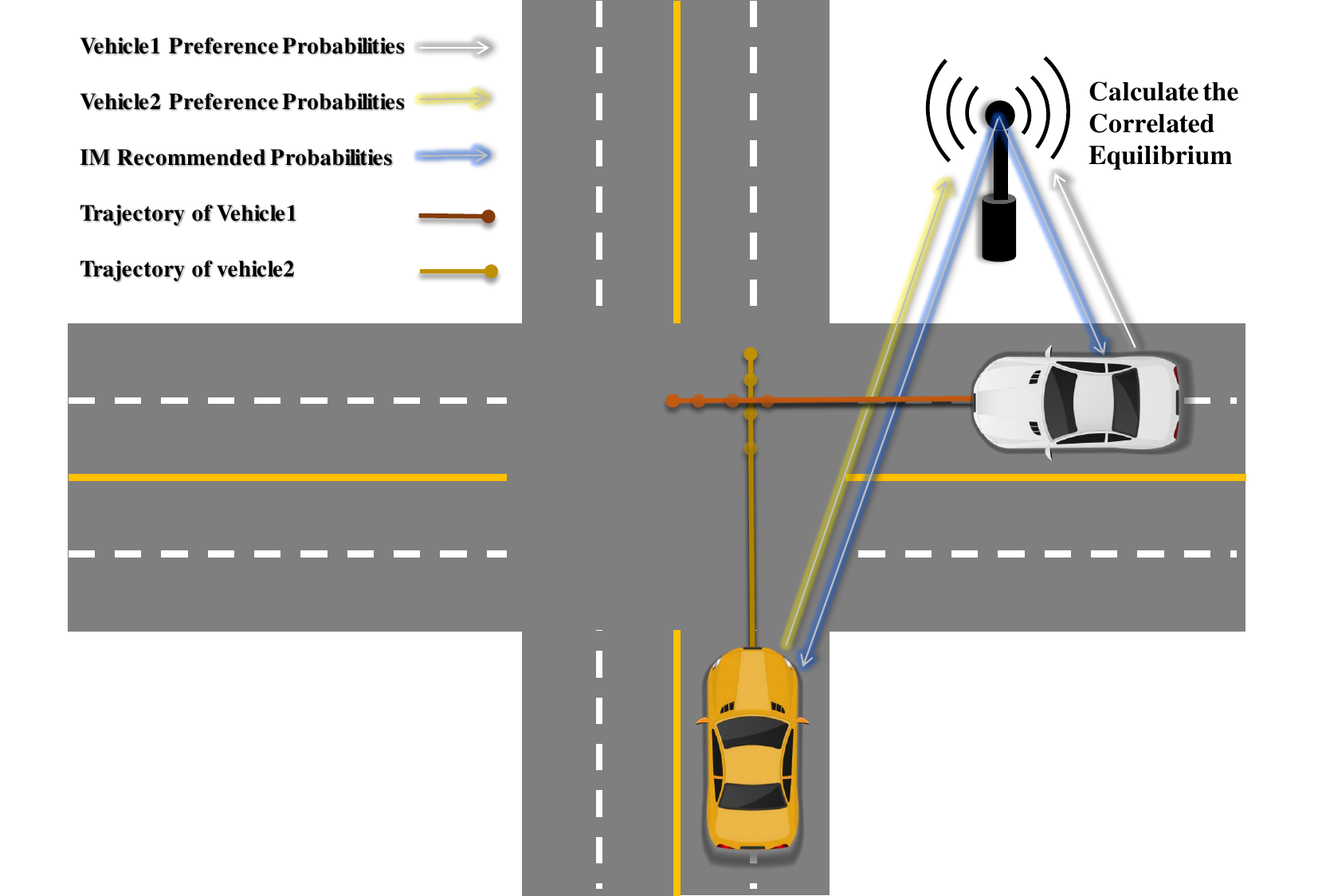}
\caption{At traffic intersections, the IM collects the preference probability distributions of vehicles over the trajectory library and computes recommendation probability with a CE constraint that aligns with the interests and rationality of all vehicles.  The experimental details are provided in Sect. VI. The video is available at \href{https://www.bilibili.com/video/BV1MZ421e7L7/?spm_id_from=333.337.search-card.all.click&vd_source=b35525d31a331a616aa4def5c50e002b}{https://www.bilibili.com/video/BV1MZ421e7L7/spm\_id\_from
\=333.337.searchcard.all.click\&vd\_source=b35525d31a331a616a
a4def5c50e002b}.}
\label{go_four}
\end{figure}

While significant research has been conducted in the field of multi-vehicle trajectory planning at intersections, striking a balance between vehicle safety and traffic efficiency remains a challenging problem. Traffic control with signals improves safety at intersections but lacks efficiency for high traffic volumes and lacks adaptivity to real-time traffic conditions \cite{gholamhosseinian2022comprehensive}. On the other hand, autonomous driving offers flexibility for optimizing vehicle behavior at intersections. Most existing vehicle trajectory generation methods adopt a predict-then-plan strategy. However, this methodology ignores the interaction among the vehicles,  leading to large inter-vehicle gaps and resulting in frozen robots phenomenon and traffic efficiency loss. Hence,  game theoretical multi-vehicle planning based on the concept of NE is proposed to enhance intersection passing through efficiency, like Tomlin et al. \cite{fridovich2020efficient}. Nevertheless, if multiple vehicles fail to converge to a consensual NE, there is a potential risk of misinterpreting opponent vehicles' intentions, leading to unsafe interactions.  
As V2I communication becomes available, autonomous vehicles approaching intersections can exchange information with infrastructure or other vehicles to arrange or coordinate their actions \cite{dresner2006human}. This enables them to avoid unnecessary stops, improve traffic efficiency, and most importantly, minimize accidents at intersections. However, this system-level optimization cannot fully account for individual rationality, and even if globally optimal strategies are computed, there is no guarantee that each vehicle will conform to them.

In this work, we propose a framework based on CE, which is an important concept in game theory, to address this challenging problem. In contrast to NE with players' actions independent of each other, CE assumes that players' strategy choices are correlated, and can effectively resolve the issue of non-uniqueness of equilibrium. In fact, CE typically involves an observer who recommends to each player the pure strategies they should play, with this recommended strategy determined by a probability distribution over the set of pure strategy vectors \cite{maschler2013game}. Additionally, the set of CE distributions is a convex polytope, allowing linear programming methods to easily find CE points \cite{nau2004geometry}.

\begin{figure}[ht!] 
\centering
\includegraphics[width=3.5in]{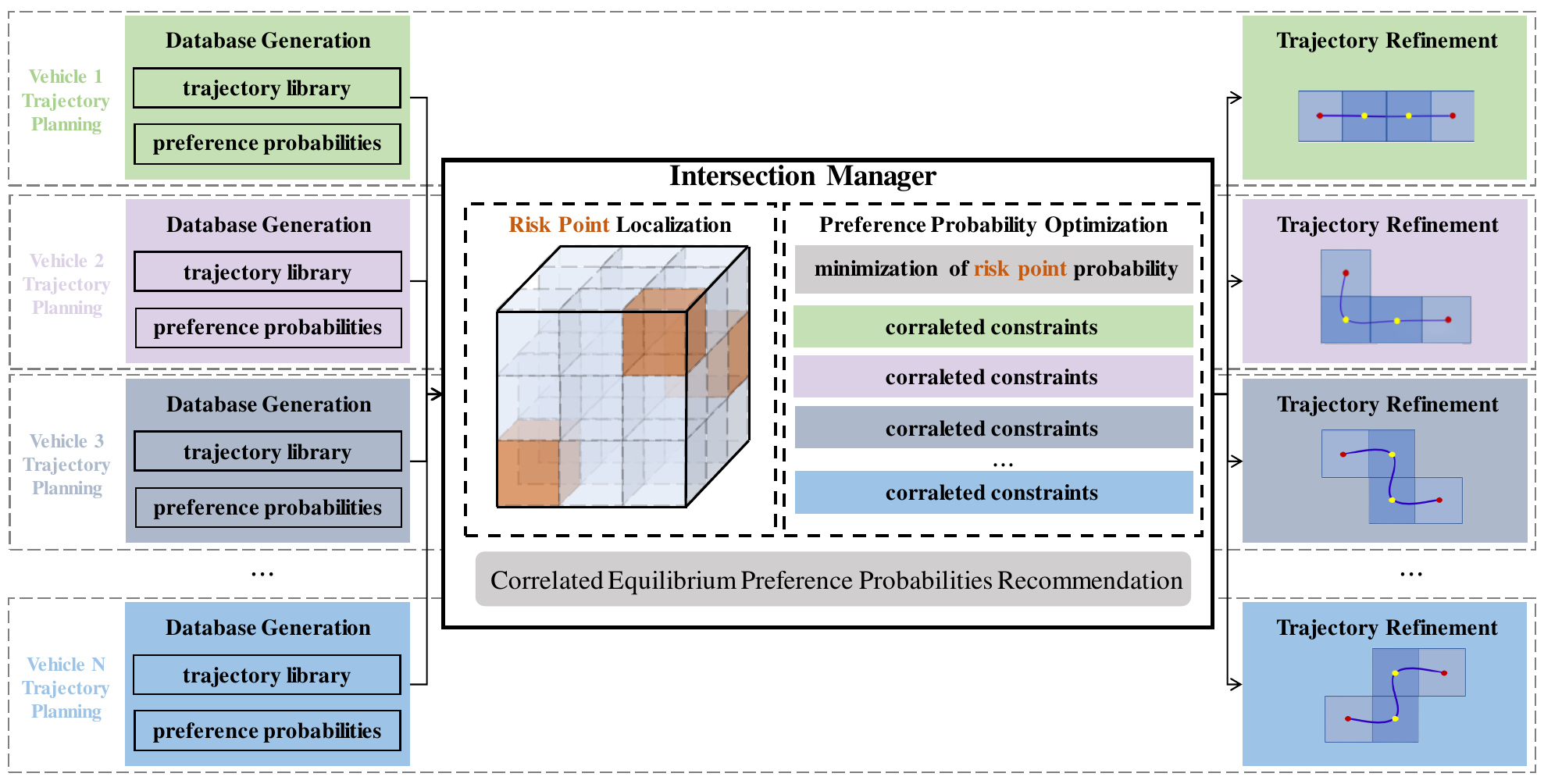}
\caption{Framework Flowchart.}
\label{framework}
\end{figure}

Within our framework shown in Figure \ref{go_four}, the IM first collects the preference probabilities of each vehicle for its trajectory library, and then optimizes the probability distribution with the constraint induced by CE, finally, the probability is recommended to the vehicles. Similar to NE, CE complies with individual rationality, ensuring that no player has an incentive to deviate from the equilibrium. Technically, each vehicle generates a trajectory library on a low-resolution spatial-temporal grid map, along with its preference probability over the trajectory library, which reflects its utility. A low-resolution spatial-temporal grid map is an extension of a two-dimensional grid map with the z-axis representing time. Afterward, the IM collects information from each vehicle and constructs an optimization problem to seek a recommendation probability for each vehicle with a CE constraint. The optimization problem maximizes the social utility by reducing the probability of collision-risky points, which are points when trajectories from different vehicle libraries coincide within the same spatial-temporal grid.  Due to the rationality of CE, each vehicle should sample a trajectory based on the probability distribution recommended by the IM. Then, by constructing a safety corridor using the low-resolution grid trajectory, each vehicle further generates a smooth and dynamic feasible trajectory through another local refinement procedure. The overall framework flow can be seen in Figure \ref{framework}.

To the best knowledge of us, the proposed approach is the first try to utilize CE for trajectory planning at traffic intersections. Compared to existing works, the proposed method effectively balances traffic efficiency and vehicle safety while maintaining computational efficiency. By comparing it with two other methods, i.e. Algames and MPC, we demonstrate the effectiveness of our approach in multi-vehicle scenarios and showcase its online operation ability. We summarize our contributions as follows:
\begin{itemize}
    \item We propose a CE-based framework for multi-vehicle trajectory planning to optimally balance traffic efficiency and vehicle safety at V2I-enabled traffic intersections.
    \item The recommendation over the low-resolution trajectories is used by each vehicle to construct safe corridors and generate refined vehicle trajectories that are both dynamics feasible and interaction-safe.
    \item We conduct hardware testing of our algorithm on laboratory miniature Ackerman cars and compare it with two other algorithms, validating the effectiveness of our approach.
\end{itemize}

\section{Related Work}

\textbf{A.} \emph{Nash Equilibrium-based Trajectory Planning}

Many existing studies used NE to define the expected interaction outcome among multi-vehicles. In \cite{wang2021game}, iterative best response (IBR) was utilized for computing NE-based trajectories. And \cite{le2022algames} introduced an augmented Lagrangian solver capable of efficiently solving generalized NE problems. The above methods incorporated the idea of receding horizon optimization from model predictive control (MPC) to effectively adapt to environmental changes, and an open-loop NE-based planning was derived. On the other hand,  in \cite{laine2019efficient} the feedback trajectory generation strategies were studied with the closed-loop NE of dynamical game.  However, the above game-based methods assumed that each vehicle independently constructs the game model and solved the NE. If the vehicles construct different game models or compute different NE, there is a risk of misjudging the intentions of interacting vehicles, potentially leading to traffic accidents. Furthermore, as the number of agents increased, the computational efficiency of such algorithms significantly decreased.

\textbf{B.} \emph{Cooperative Resource Reservation in  conflicts resolution }
 
Many studies decomposed intersections in terms of time and space, generated low-resolution spatial-temporal grid maps, and then employed cooperative resource reservation to resolve conflicts. Depending on the involvement of infrastructure, cooperative resource reservation could be classified as centralization and distribution \cite{7244203}. For centralized algorithms, the allocation of spatial-temporal resources was primarily managed by the infrastructure. For instance, \cite{dresner2004multiagent} assumed that once a grid reservation was approved, vehicles maintained the same speed until they passed through the intersection.  \cite{dresner2006human} considered emergency vehicles and proposed a traffic light-based resource reservation system. In the case of distributed algorithms, all resource allocations were negotiated among the vehicles themselves. In \cite{azimi2011vehicular,azimi2012intersection}, a set of distributed reservation protocols based on message communication was proposed for intersections and roundabouts. In \cite{azimi2011vehicular}, the reservation and release of spatial-temporal resources were managed through Stop and Clear messages. In a later work \cite{azimi2012intersection}, considering vehicle dynamics such as acceleration and deceleration, additional Input messages were introduced to optimize the management of spatial-temporal resources.

\section{Vehicle Trajectory Library Generation}

In this section, we present a method for generating a trajectory library in low-resolution spatial-temporal grid maps and the associated preference probabilities over the library for each vehicle. Specifically, we first utilize the A* algorithm to generate a feasible path, and then employ time allocation techniques to generate trajectory libraries for each vehicle. Each vehicle decides its own preference probabilities for each trajectory, by using the multinomial logit (MNL) choice model with personalized parameters. The vehicles can prioritize quick passage through intersections, or prioritize safety considerations. The library and preference probability are collected by the IM for the computation of CE-based recommendations.

 \textbf{A.} \emph{Trajectory Library Generation}
 
 Each vehicle generates a path using the A* approach, denoted as $\{x^i_k\}_{k\in [0,L]}$, where $x_k$ is the coordinate in the Cartesian coordinate system, and the size of a grid is larger than the size of the vehicle. 

\begin{figure}[ht!] 
\centering
\includegraphics[width=3.5in]{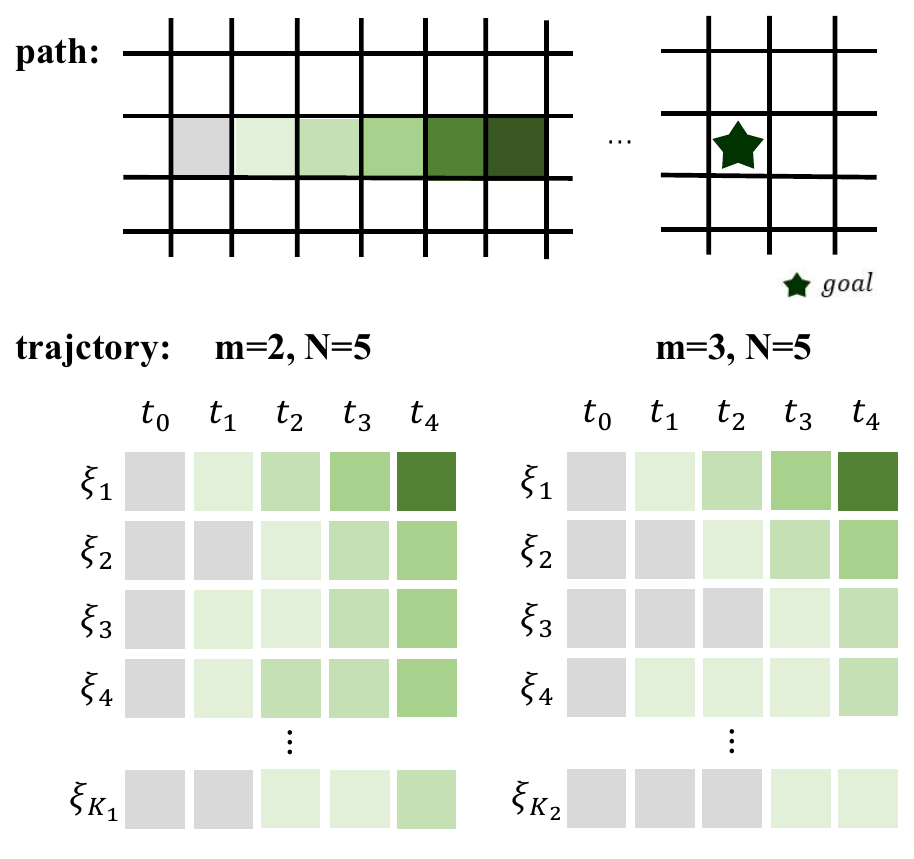}
\caption{Trajectory generation diagram.}
\label{traj_gene}
\end{figure}

At the current time $t_0$, the projected point of the vehicle on the path serves as the starting point and is indexed as $l$. Each vehicle stays at each path point at most $m$ time steps within a total planning horizon of $N$ time steps. The generated trajectories with different time allocations are shown in Figure  \ref{traj_gene}. In this figure, the path represents the sequence of points obtained through A* search, with a darker color indicating closer proximity to the destination. For the case of $m=2$ and $N=5$, each colored point has the possibility of appearing consecutively for two-time steps. Similarly, for $m=3$ and $N=5$, each point can appear consecutively for a maximum of three time steps. Here we use $\xi$ to represent a trajectory and use $\boldsymbol{\xi}$ to represent a library of trajectories, the pseudo-code for the trajectory generating algorithm is shown in Algorithm \ref{alg1}.

\begin{algorithm}
\renewcommand{\algorithmicrequire}{\textbf{Input:}}
	\renewcommand{\algorithmicensure}{\textbf{Output:}}
    \caption{GenerateTrajectoryLibrary}
    \label{alg1}
    \begin{algorithmic}
            \REQUIRE initial path $\{p_k\}_{k\in [0,L]}$, time horizon $N$, length of maximal stay $m$, index $l$
            \ENSURE trajectories library $\boldsymbol{\xi}$
            \FOR{$i = 0,1,...,m-1$}
            {
            \STATE $\xi$.push$\_$back$(p_l)$;
            \IF{$\xi$.size = $N$}
            \STATE $\boldsymbol{\xi}$.push$\_$back$(\xi)$;
            \FOR{$j = 0,1,...,i$}
            \STATE $\xi$.pop$\_$back();
            \ENDFOR
            \RETURN 
            \ELSE 
            \STATE GenerateTrajectoryLibrary$(\{p_k\}_{k\in [0,L]},N,m,l + 1)$;
            \ENDIF
            }
            \ENDFOR
            \STATE $\xi$.push$\_$back$(p_l)$;
            \IF{$\xi$.size() = $N$}
            \STATE $\boldsymbol{\xi}$.push$\_$back$(\xi)$;
            \ELSE
            \STATE GenerateTrajectoryLibrary$(\{p_k\}_{k\in [0,L]},N,m,l + 1)$;
            \ENDIF
            \FOR{$i = 0,1,...,m-1$}
            \STATE $\xi$.pop\_back();
            \ENDFOR
            \RETURN 
    \end{algorithmic} 
\end{algorithm}

 \textbf{B.} \emph{Preference Probabilities Calculation}

Each vehicle has its preference for the generated trajectory set. Denote $P_{n,o}^i$ as the initial preference of vehicle $i$ for trajectory $n$. Here a preference of each trajectory is computed based on its comfort (average acceleration) and efficiency. With a fixed time interval between trajectory points, the acceleration is directly calculated from the second-order difference of positions, and the efficiency is measured by the trajectory length $s^i_n$. Then according to $P_{n,o}^i$, the initial preference probability $p^i_{n,o}$ that vehicle $i$ places on the trajectory $\xi^i_n$  could be calculated  by a MNL choice model:

\begin{align}
p^i_{n,o} = \frac{e^{V^i_n}}{\sum_{n=1}^{k_i}{e^{V^i_n}}},\label{probability constraint}
\end{align}
where $V^i_n = -(\alpha^{i}+\beta^{i}P_{n,o}^i)$ and $\alpha^{i}$,$\beta^{i}$ are vehicle-specific constants measuring the sensitivity of preference probabilities.


\section{ IM recommendation with Correlated Equilibrium}

In this section, the IM optimizes welfare maximizing preferences in light of CE, accepted by each vehicle. The IM gathers information from each vehicle and establishes a CE constraint for each vehicle. 
The CE constraint accounts for its local passing through efficiency and safety and also favors a probability over a library with a low entropy \cite{ning2023robust}. The IM objective function is the overall system safety represented by the sum of probabilities for all risk points, which are grid cells in the spatial-temporal grid map where multiple vehicles' trajectories intersect.  

\textbf{A.} \emph{Risk Point Localization}

\begin{figure}[ht!] 
\centering
\includegraphics[width=3in]{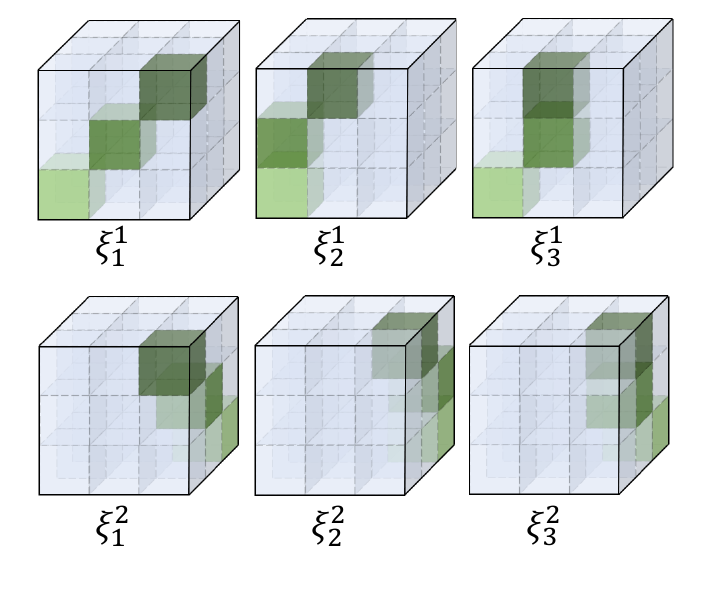}
\caption{Representation of trajectory library in spatial-temporal grid map.}
\label{sep_tem_traj_demo}
\end{figure}

Iterating over all trajectories from different vehicles' trajectory libraries for collision detection to identify all risk points is time-consuming. By constructing a key-value pair for the map, we can find all collision points in $O(n)$ time complexity. The key is each grid in the spatial-temporal grid map, and the value is the set of vehicle IDs passing through that grid. When a vehicle's trajectory passes through a grid (key), the corresponding value of that grid (key) is incremented by one. In this way, each trajectory can quickly find its corresponding risk point, reducing collision detection time. Taking an intersection as an example, for convenience, take $m=2$ and $N=3$. As shown in Figure \ref{sep_tem_traj_demo}, the trajectory library generated by the first vehicle, which intends to move from left to right, is represented in the spatial-temporal grid map by the top three cubes, while the trajectory library by the second vehicle, moving from back to front, is represented by the bottom three cubes. If there are a total of $I$ participants at the intersection, $\xi^i_n$ denotes the $n$-th trajectory of the $i$-th participant, where $i = 1,2,...,I$ and $n = 1,2,...,k_i$.
  
\begin{figure}[ht!] 
\centering
\includegraphics[width=3in]{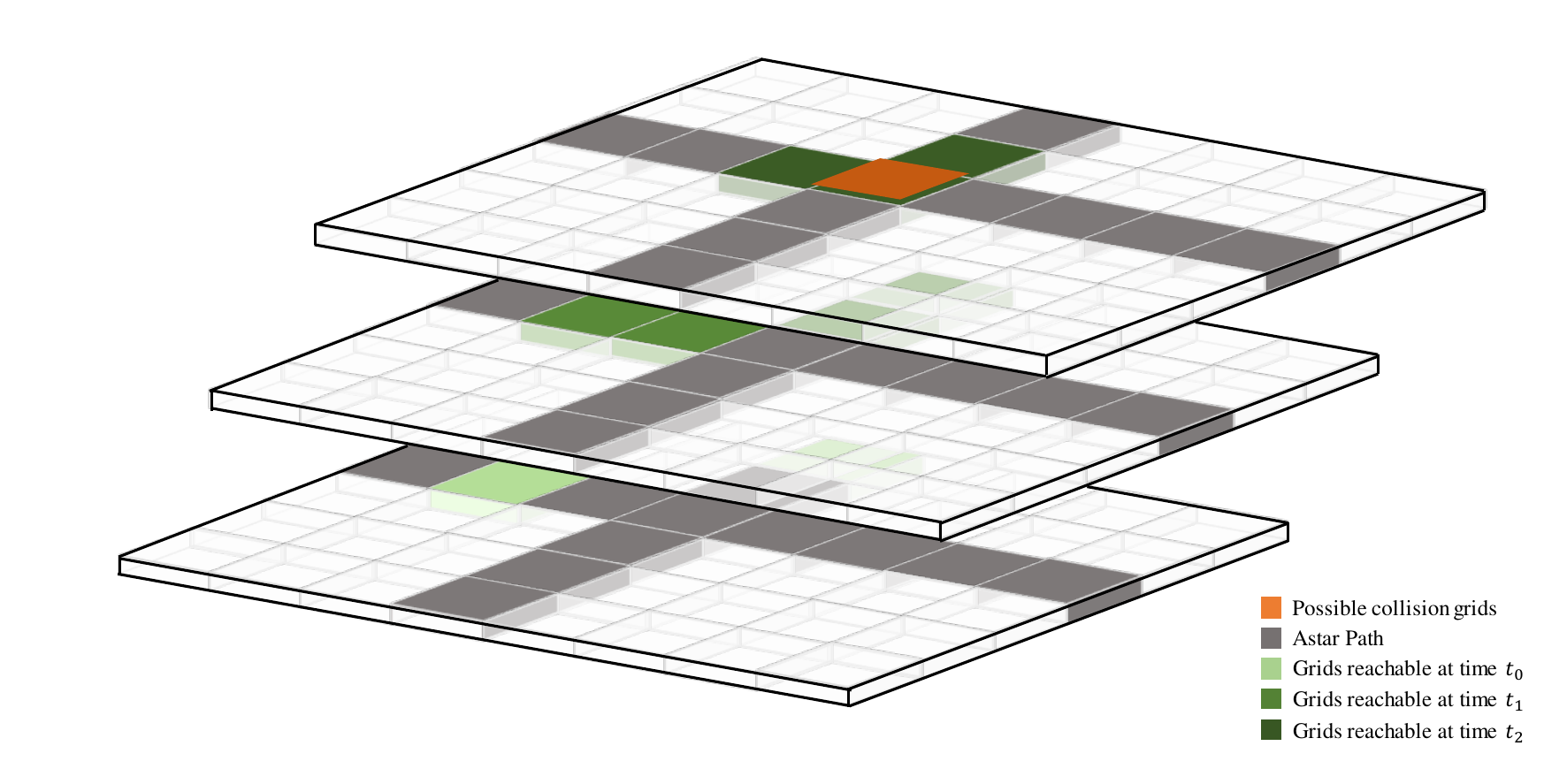}
\caption{The schematic diagram of information is contained in the spatial-temporal grid map.}
\label{map}
\end{figure}

For each additional vehicle, we make a corresponding adjustment to the number of vehicles (value) corresponding to each raster (key) in the trajectory. Figure \ref{map} includes information such as the paths of each vehicle (in gray), the trajectory libraries (in green), and risk points between trajectory libraries (in orange). Here, only the orange point corresponds to vehicles $1$ and $2$, while the other green points correspond to either vehicle $1$ or vehicle $2$.

\textbf{B.} \emph{Correlated Equilibrium Constraints Construction}

We first briefly introduce the concept of CE and then specify its usage in the proposed framework. We consider an $N$-player finite game, each player $j$ has a finite set of strategies $S_j$. $\mathcal{S}=\prod \limits_{j=1}^NS_j$ is the set of strategy profiles. We denote $\prod \limits_{k\neq j}S_k$ by $S_{-j}$. The utility function of player $j$ is a function $u^j$ mapping $\mathcal{S}$ to a number. A distribution $\mathcal{P}$ on $\mathcal{S}$ is a product if for each player $j$ there is a distribution $\mathcal{P}_j$ on $S_j$ such that for all $s=(s_1,...,s_N)\in \mathcal{S}$, $\mathcal{P}_s = \prod \limits_{j=1}^np_{s_j}^j$. A CE is a distribution $\mathcal{P}$ on $\mathcal{S}$ such that for all players $j$ and all strategies $s_j^k,s_j^l\in S_j$, the following inequality is satisfied:
\begin{align}
\sum_{s_{-j}\in S_{-j}}[u^j(s_j^k,s_{-j})-u^j(s_j^l,s_{-j})]\mathcal{P}(s_j^k,s_{-j})\geq 0, \forall j\label{ce_defin}
\end{align}

Similar to \cite{ning2023robust}, we consider there are only two alternatives for each vehicle in their strategies set $S_j$, i.e., $S_j = \{s_j^1,s_j^2\}$. Strategy $s_j^1$ means that the vehicle $j$ follows its individual selfish initial preference probabilities $\{p^j_{n,o}\},n=1,...,k_j$, and $s_j^2$ means that the vehicle $j$ follows the suggested preference probabilities $\{p^j_n\},n=1,...,k_j$ given by the IM. Denote the strategy profile that all vehicles follow the suggested preference probabilities(take the strategy $s_j^2$) as $s_{f} = \{s_1^2,...,s_N^2\}$. We aim to achieve CE, satisfying the condition of $\mathcal{P}_{s_f} = 1$ and $\mathcal{P}_s = 0,\forall s\neq s_{f}$. Then Eq.\ref{ce_defin} becomes:
\begin{align}
0 + 1*(u^j(s_j^2,s_{-j}^2)-u^j(s_j^1,s_{-j}^2))\geq0, \forall j\label{ce_con}
\end{align}
where $s_{-j}^2$ represents every vehicle other than $j$ takes the strategy $s_j^2$. For every rational vehicle, as long as it satisfies the CE constraint, it has no reason to maintain its initial preference probabilities. In other words,  the choices provided by the IM, which has access to global information, are superior to the choices generated by each vehicle based on local information.

\textbf{C.} \emph{Preference Probability Optimization}

As the probability satisfying correlation constraints may result in unsafe interaction, we aim to identify optimal recommended strategies that minimize the probability of collisions at the intersection, thereby enhancing social utility. From the spatial-temporal grid map, we can obtain the number and locations of risk points. Denote $I_m$ as the set of participants at the $m$-th risk point in the map and $m = 1, 2, ..., M$, where $M$ is the total number of risk points in the entire map, and $I_m^i$ as the set of collision trajectories that vehicle $i$ passes through at the $m$-th risk point. For vehicle $i$, its probability at the risk point is:
 \begin{align}
p^{i,m} = \sum_{n\in I_m^i}p_n^i,\label{prob_i_m}
\end{align}
and the probability of collision occurrence at the risk point is:
 \begin{align}
p^m = \prod \limits_{i\in I_m}p^{i,m}.\label{prob_m}
\end{align}

Then the IM recommends optimal trajectory preferences for individual vehicles by solving the following problem:
 \begin{align}
\min_{p_1^1,...,p_{k_I}^I}\sum_{m=1}^M(p^m)\label{opt_1}
\end{align}
\begin{align}
s.t.&\quad\sum_{n=1}^{k_i}\left(p_n^ilogp_n^i+p_n^is_n^i\right)-\sum_{V^i}\left(p^{i,m}_c\prod_{j\in I_m\setminus\{i\}}p^{j,m}\right)\notag\\&\quad\geq\sum_{n=1}^{k_i}\left(p_{n,o}^ilogp_{n,o}^i+p_{n,o}^is_n^i\right)-\sum_{V^i}\left(p^{i,m}_o\prod_{j\in I_m\setminus\{i\}}p^{j,m}\right),\notag\\&\quad\forall i= 1,...,I \label{opt_2}
\\&\quad \sum_{n=1}^{k_i}p_n^i=1,\forall i= 1,...,I \label{opt_3}
\\&\quad p_n^i\geq\epsilon,\forall i= 1,...,I,\forall n = 1,..,k_i\label{opt_4}
\end{align}
where (\ref{opt_1}) minimizes the probability of collision occurrences and (\ref{opt_2}) represents the CE constraint. The CE constraint in (\ref{opt_2}) incorporates three criteria: entropy of recommendation probability, expected efficiency of each vehicle, and the probability of collision occurrences for each vehicle. Particularly, define $V_i$ as the set of risk points for vehicle $i$, and 
\begin{align}
p^{i,m}_c =
\begin{cases}
1- p^{i,m} \quad d \leq d_{tor} \\
p^{i,m} \quad else
\end{cases} ,\label{def_c}
\end{align}
where $d$ represents the grid distance of the vehicle to the risk point, $d_{tor}$ is a threshold. This formula indicates a preference for vehicles to pass first when they are near a risk point, and vice versa.

We can utilize the Ipopt \cite{wachter2006implementation} solver for the optimization problem. Each vehicle selects a trajectory based on the optimized preference probability distribution.

\section{Trajectory Refinement}

\begin{figure}[ht!] 
\centering
\includegraphics[width=3.5in]{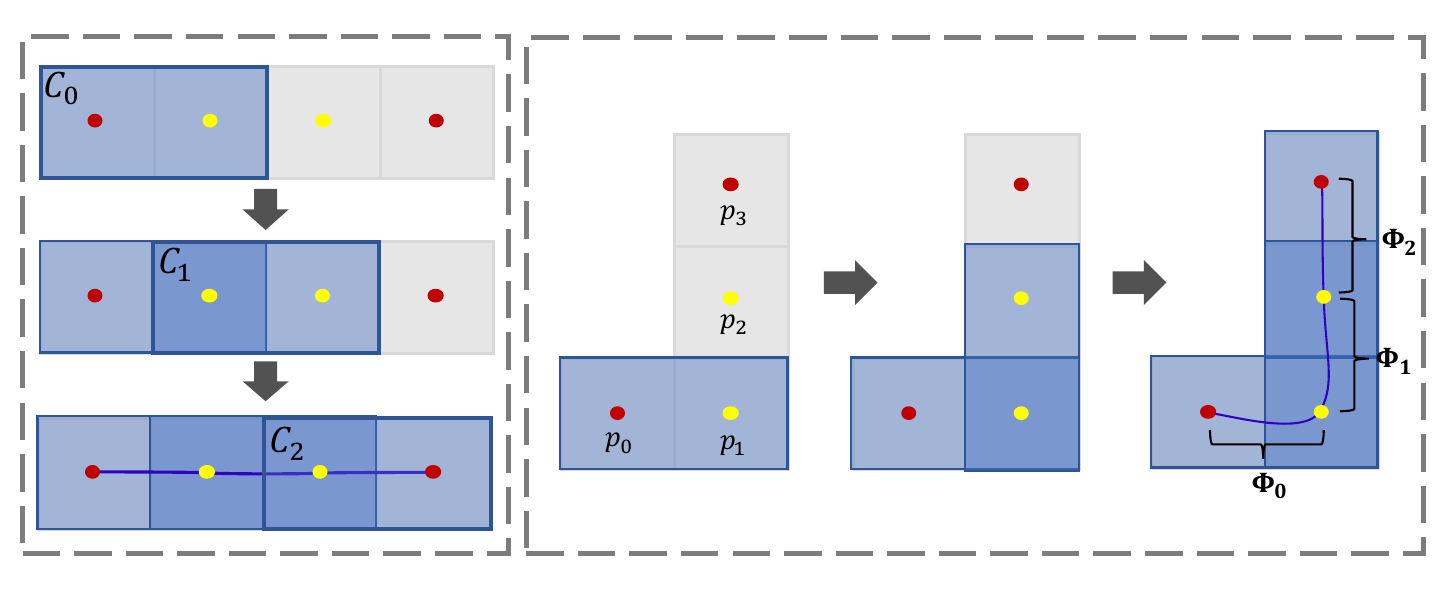}
\caption{The schematic diagram illustrating the construction of the safety corridor. }
\label{corridor}
\end{figure}

The resolution of the spatial-temporal grid map is relatively low, with each grid unit equal to or larger than the size of a vehicle. Hence, the coarse trajectory lacks dynamical feasibility and cannot be tracked by the vehicle controller, so we need to refine the trajectories. Adopt the coarse grid trajectories sampled by each vehicle to construct a safety corridor. The construction of the safety corridor is illustrated in Figure \ref{corridor},  where the red dots indicate the starting and ending points, and  the yellow dots represent the waypoints.  We take two consecutive trajectory points in time, where each point represents a grid cell. As a result, these two points form a rectangle. We denote $i$-th rectangle as a convex space $C_i$, and the trajectory between the two points is represented by a polynomial function $\Phi_i$. Once the coarse grid trajectory can generate $n$ rectangles, the entire trajectory consists of $n$ polynomial functions, with the $i$-th polynomial function located within the $i$-th rectangle. If the trajectory points are the same, they are skipped. 

We utilize the methodology described in \cite{liu2017planning} to optimize the trajectories. The convex optimization can be formed as a QP with constraints as  

 \begin{align}
\arg\min_{\Phi}\sum_{i=1}^n(\int_{0}^{\Delta t_i}  ||\frac{d^3}{dt^3}\Phi_i(t)||^2 dt)\label{qp_1}
\end{align}
\begin{align}
s.t. &\quad\frac{d^k}{dt^k}\Phi_i(\Delta t_i) = \frac{d^k}{dt^k}\Phi_{i+1}(0), k=0,...,3 \label{qp_2}
\\&\quad A^T_i\Phi_i(t)\textless b_i ,\forall i= 1,...,n \label{qp_3}
\end{align}
Here \ref{qp_3} is a polygonal constraint, $A_i$ and $b_i$ correspond to the $i$-th polyhedron $C_i$. $\Phi_i(t)$, $\frac{d}{dt}\Phi_i(t)$, $\frac{d^2}{dt^2}\Phi_i(t)$, $\frac{d^3}{dt^3}\Phi_i(t)$ indicate the desired position, velocity, acceleration and jerk at time $t$. $\Delta t_i$ in above equations refers to time of each polynomial as $\Delta t_i = t_{i+1} - t_i$, which is equal to the temporal resolution.


\section{Experiments}
In this section, we present the experimental results of the algorithm in various scenarios and compare its performance with that of the game-based algorithm Algames \cite{cleac2019algames}, as well as the traditional non-game-based MPC approach \cite{fan2018baidu}.

\textbf{A.} \emph{Experimental Setup}

\begin{figure}[ht!] 
\centering
\includegraphics[width=3.5in]{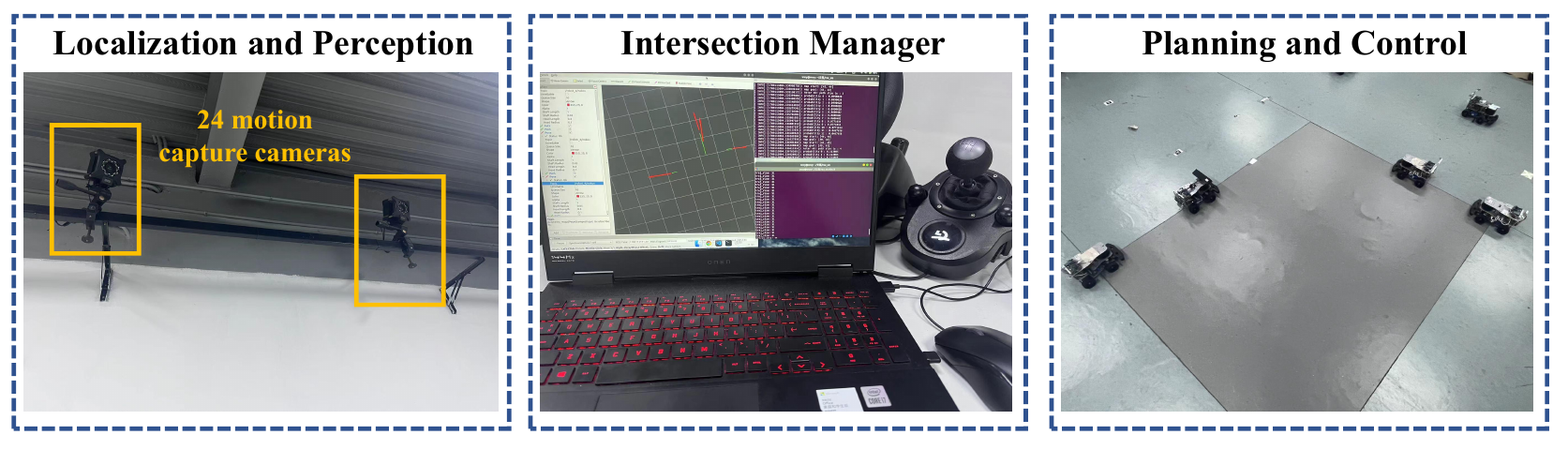}
\caption{Experimental framework for a motion capture system and Ackermann vehicle.}
\label{exp_framework}
\end{figure}

1) \emph{Hardware Platform: }We deploy our algorithm on Ackermann miniature vehicle models. The system architecture, shown in Figure \ref{exp_framework}, employs a motion capture system for precise localization and perception. Data transmission is facilitated through a 5G network, while the IM features an Intel Core i7 10750H processor with 16GB of memory. The Planning and Control modules are executed on an onboard Nvidia TX1 computer.

2) \emph{Implementation Details: }We implement our algorithm in C++ within the Robot Operating System (ROS) framework. Set $m = 2$, $N = 6$ for library generation, $\alpha^i=0$, $\beta^i=1$ for initial preference initialization, $d_{tor}=2$ for optimization in all experiments. We set the spatial resolution of the spatial-temporal grid map to 0.3 meters, which is equal to the length of a vehicle, and the temporal resolution to 0.6 meters.

3) \emph{Evaluation Metrics: }Choose four evaluation metrics to compare with the other two algorithms: planning execution frequency $f$, total transit time $t_{total}$, minimum distance between vehicles in different directions $d_{min}$, and the weight of focus on safety $w_{s}$ and efficiency $w_{e}$. The weight calculation process is as follows: we first perform min-max normalization on the $d_{min}$ and the inverse of the $t_{total}$ of three algorithms separately, and then apply softmax to the normalized data of each algorithm to derive the respective weights.

4) \emph{Receding Horizon Planning: }To address the challenge of long distances between the starting and ending points of vehicles, generating the entire trajectory library at once using the algorithm described in Chapter III would require significant computational resources and time. To improve efficiency, we adopt a receding-horizon planning scheme. We perform global path planning only once, considering that it is reasonable for structured road environments like intersections.  We only use a portion of this path to generate the trajectory library. The algorithm for generating the trajectory library selects the starting point as the projection point of the vehicle onto the global path. The vehicle only executes a small portion of the trajectory, after which the trajectory library generation process is repeated. This approach saves computational resources and reduces the time required for trajectory generation. 

\textbf{B.} \emph{Intersection Simulations}

\begin{figure}[ht!] 
\centering
\includegraphics[width=3.5in]{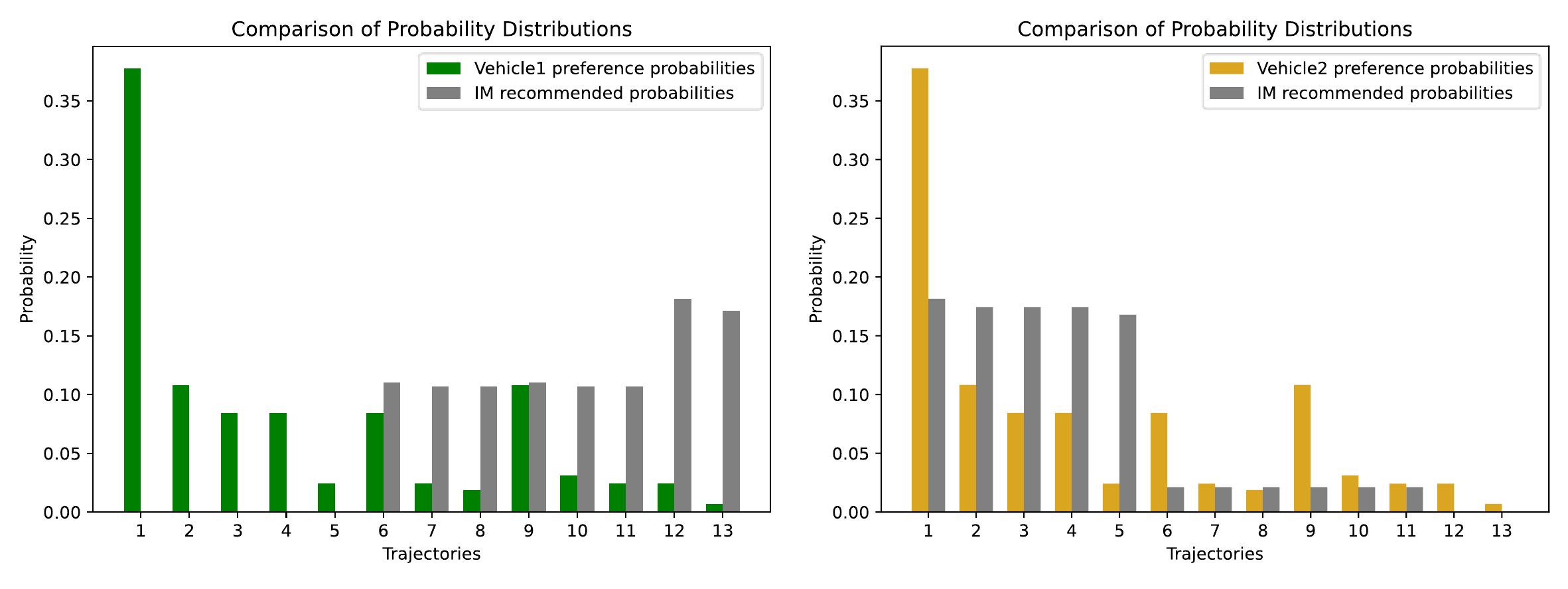}
\caption{The preference probabilities of two vehicles and the recommended probabilities calculated through the IM.}
\label{sim_prob_dist}
\end{figure}

\begin{figure}[ht!] 
\centering
\includegraphics[width=3.5in]{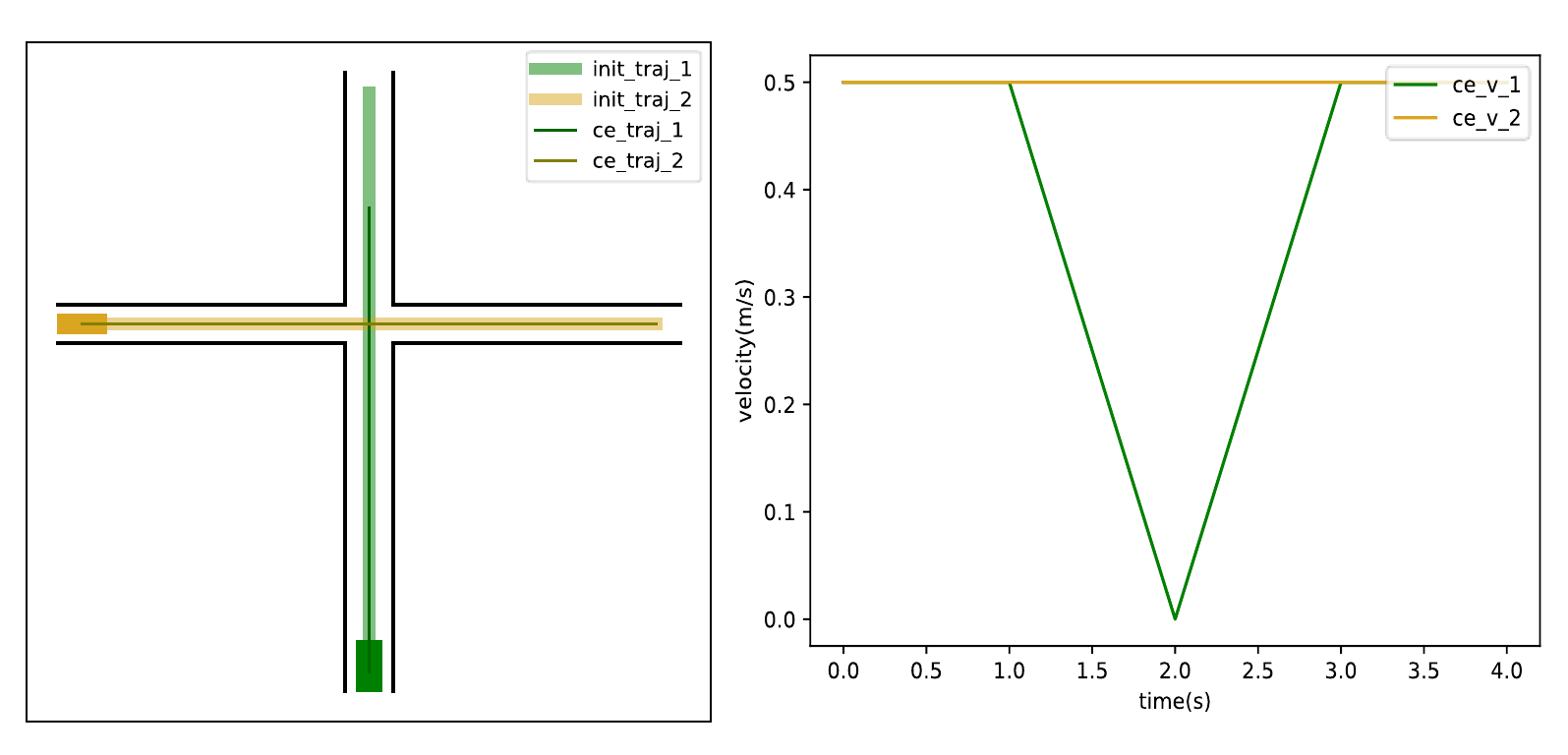}
\caption{The allocation of trajectory situations using CE in the IM.}
\label{sim_exp}
\end{figure}

We first conduct simulation experiments to validate the feasibility of our algorithm. As shown in Figure \ref{sim_prob_dist}, at the intersection, each vehicle has 13 trajectories to choose from.
The green and gold colors represent the original preference probabilities of two vehicles over the trajectories, while the gray color represents the preference probabilities recommended by the IM. Following the recommended probability distribution by the IM, it is ensured that the second vehicle passes prioritized while the first vehicle yields and stops. This guarantees a smooth passage through the intersection for both vehicles. Figure \ref{sim_exp} represents one possible scenario of trajectory selection based on the recommended probability distribution. The left image depicts the vehicles' trajectories before and after the IM computation, while the right image illustrates the recommended trajectory speeds. From the left image of Figure \ref{sim_exp}, we can observe that both vehicles initially preferred a straight-line trajectory (represented by bold lines). However, following their original plans may result in a collision. The IM resolves this conflict to recommend trajectories. It enables one vehicle to pass through the intersection at a consistent speed, while the other vehicle briefly pauses for a single time interval within a grid (braking once) before resuming its forward movement. 

\begin{figure}[ht!] 
\centering
\includegraphics[width=3.5in]{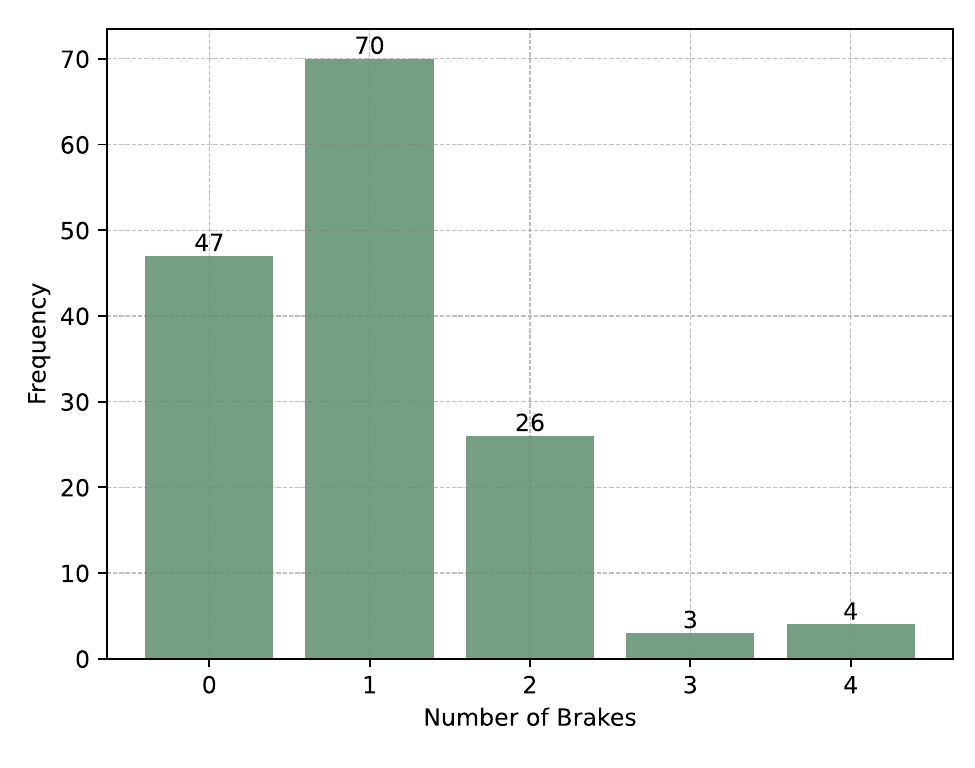}
\caption{Visual representation of the number of times two vehicles come to a complete stop in various scenarios during a Monte Carlo experiment. }
\label{sim_bar}
\end{figure}

We also conduct a Monte Carlo experiment by introducing Gaussian noise to the initial preference parameters $\alpha^i$ and $\beta^i$ provided by the users. The experiment is performed independently for 150 iterations. Throughout the experiments, we maintain the same starting point, and we assumed that vehicles traveling towards the right side of the intersection would be positioned closer to it (at a grid resolution of one cell) compared to vehicles traveling upwards. Figure \ref{sim_bar} displays a bar chart that illustrates the distribution of braking occurrences for the two vehicles. The horizontal axis represents the sum of the braking occurrences for both vehicles, while the vertical axis represents the frequency of these occurrences. Analyzing the chart based on different user preferences, we can observe the following: In approximately 31.33\% of cases, the IM suggests that both vehicles can proceed through the intersection at a constant speed without any need for braking. In approximately 64\% of cases, the vehicles only need to brake once or twice to successfully navigate the intersection. Only in about 4.67\% of cases, frequent braking is necessary. These results indicate that the IM is effective in minimizing braking instances and promoting smooth traffic flow in the majority of scenarios.

\textbf{C.} \emph{Two-Vehicle Intersection Experiments}

\begin{figure}[ht!] 
\centering
\includegraphics[width=3.5in]{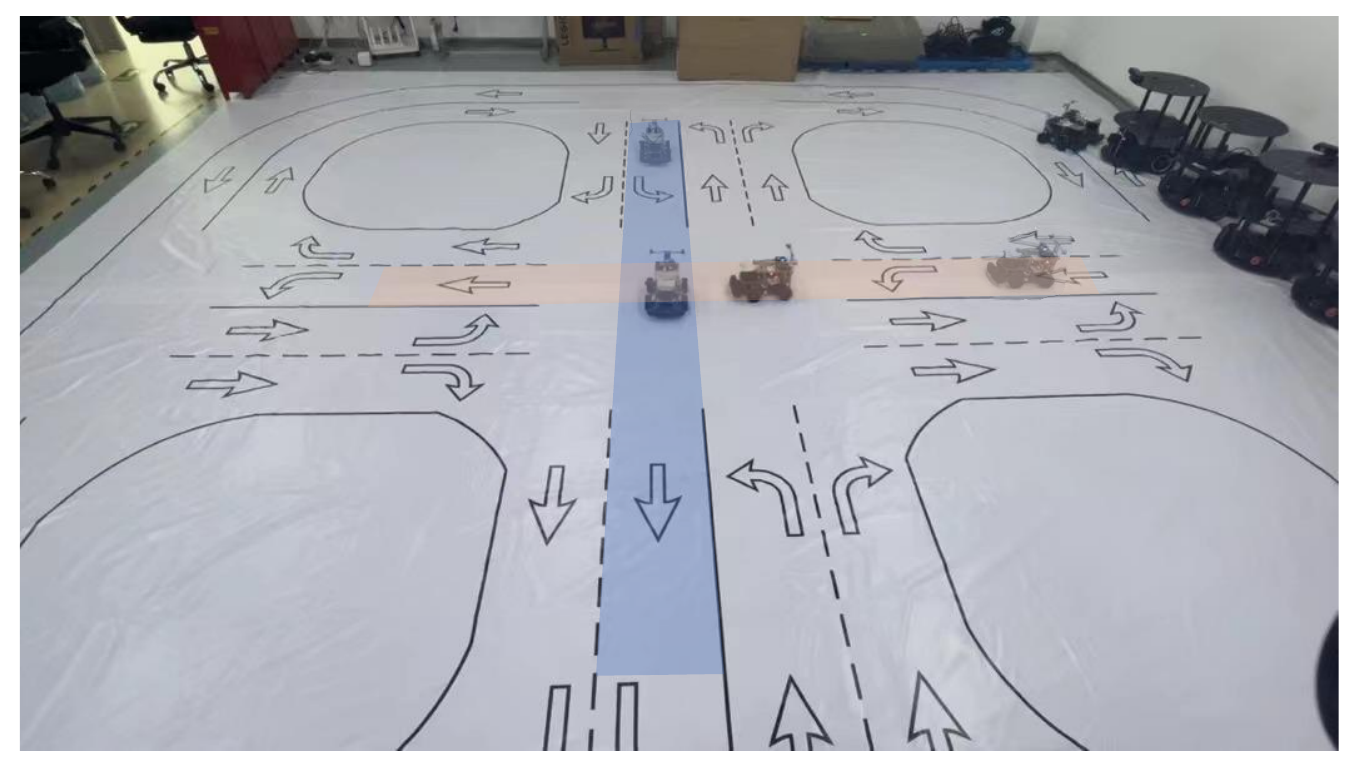}
\caption{The IM guides two vehicles through the intersection.}
\label{two_car_exp}
\end{figure}

\begin{figure}[ht!] 
\centering
\includegraphics[width=3.5in]{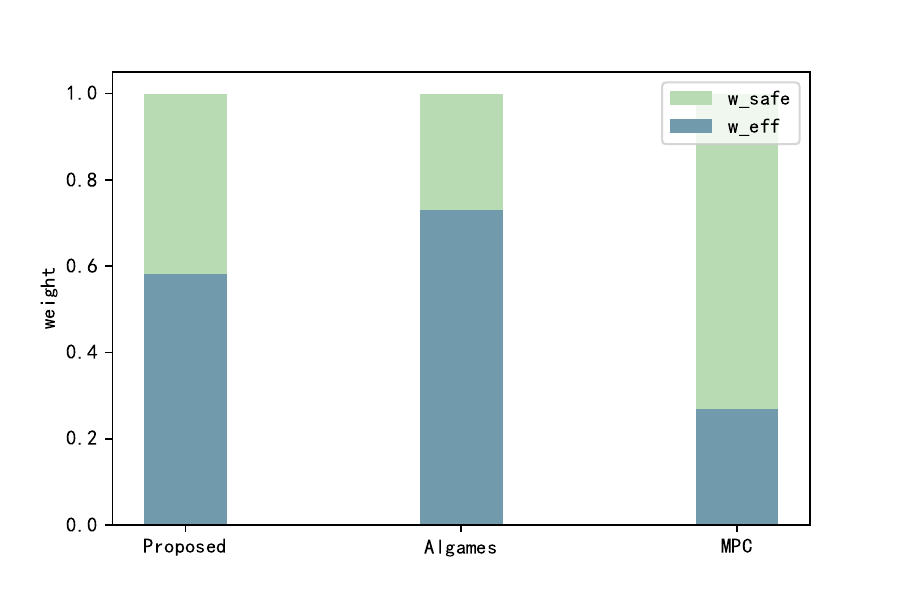}
\caption{Bar chart of the weights of safety and efficiency focus for different algorithms.}
\label{weight_three}
\end{figure}

We carry out a crossroad experiment involving two vehicles as illustrated in Figure \ref{two_car_exp}, 
and compared our algorithm  with the Algames and MPC methods with  detailed data in Table \ref{two car compare}.  The MPC method tends to adopt a conservative approach, with one vehicle waiting until the other vehicle has completely passed, resulting in a significant safety distance and consuming substantial time-space resources. On the other hand, the NE-based Algames algorithm exhibits the opposite behavior of MPC. The two vehicles compete for time-space resources and pass with a smaller distance between them. Our proposed algorithm strikes a balance by providing a safer distance while avoiding the overly conservative behavior of MPC. We plot the weights of safety and efficiency focus for different algorithms, as shown in Figure \ref{weight_three}. Under the same initial and final conditions, Algames spends more time focusing on transit efficiency, whereas the MPC algorithm allocates more attention to safety. Our algorithm strives to balance these two weights. 

For MPC algorithms, vehicles need to probe and guess their opponent's behavior in real time, resulting in prolonged interaction time. However, we utilize communication to reduce the uncertainty of intentions, leading to a reduction of at least 32.3\% in interaction time. For the Algames algorithm, although vehicles do not communicate with each other, they assume that all vehicles are in the same NE where no one deviates, leading to more aggressive vehicle behaviors. By optimizing collision probabilities through the IM aggregation, we increase the safety distance by 19.44\%.

\begin{table}[ht]
\caption{Comparing Two Vehicles at an Intersection}
\label{two car compare}
\begin{center}
\begin{tabular}{cccc}
\toprule
& Our  & Algames & MPC \\

\midrule
$f(hz)$ & 5.37 & 5.87 & 5.00  \\
$t_{total}(s)$ & 11.24 & 7.61 & 15.73\\
$d_{min}(m)$ & 0.43 & 0.36 & 1.91 \\

\bottomrule
\end{tabular}
\end{center}
\end{table}

\textbf{D.} \emph{Four-Vehicle Intersection Experiments}

\begin{figure}[ht!] 
\centering
\includegraphics[width=3.5in]{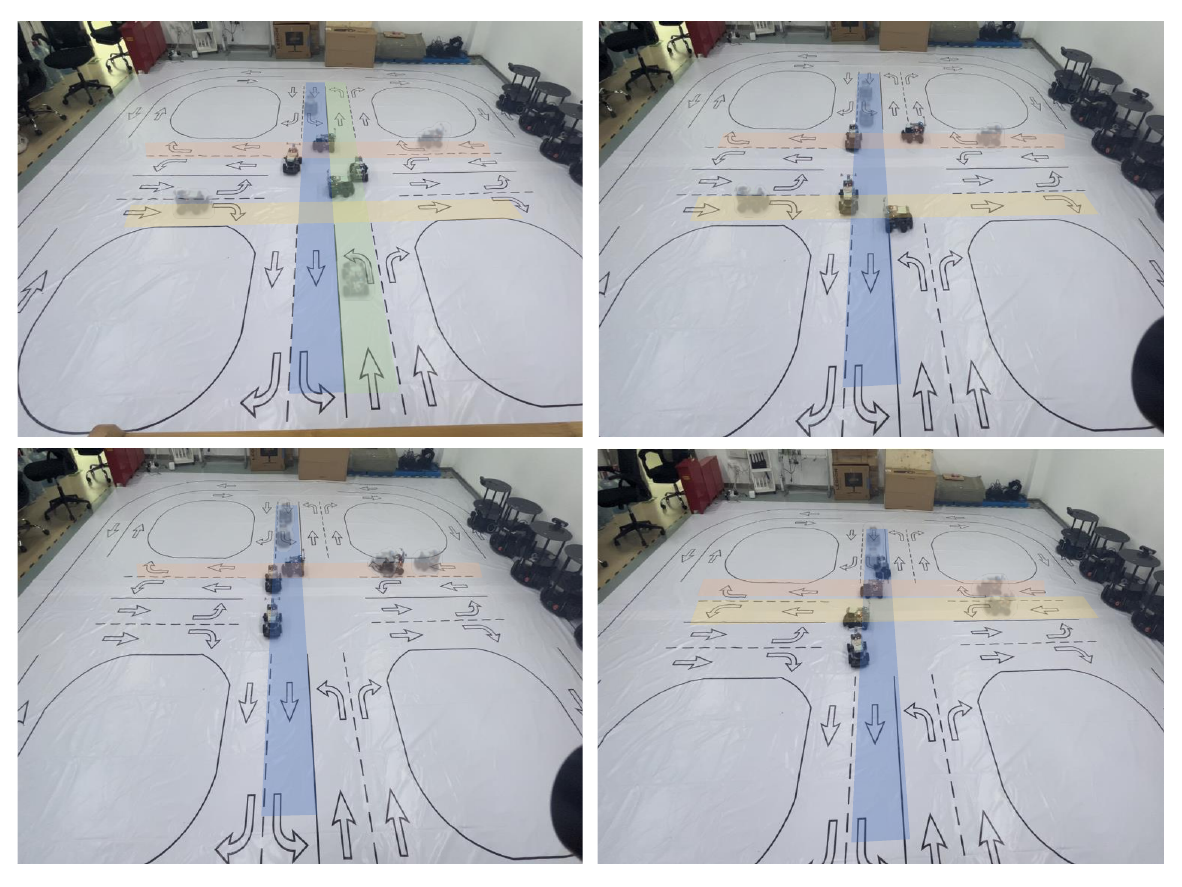}
\caption{The four different initial state scenarios of the crossroad intersection.}
\label{four_secnario}
\end{figure}

We also conduct experiments on a four-vehicle crossroad scenario, where we design four different scenarios as illustrated in Figure \ref{four_secnario}. Due to the occurrence of dimensionality explosion in Algames (computational frequency lower than 1Hz), we exclusively employ the MPC method for comparative analysis. The MPC method decouples prediction from planning, making the planning computation frequency less sensitive to the number of interacting vehicles. On the other hand, the Algames algorithm experiences a decrease in efficiency of more than 100\% due to the need to predict the trajectories of other vehicles simultaneously with planning, which hampers its real-time implementation. In contrast, our algorithm addresses the issue of dimensionality explosion by reducing the probability of collision occurrence instead of performing collision detection for each vehicle. This approach results in a moderate decrease in computational consumption, with a maximum reduction of 33.15\%.

MPC's strategy is to let vehicles pass through the intersection one by one, waiting for a vehicle to exit before another vehicle enters. In contrast, our method is more efficient. While ensuring safety, we achieve an average reduction of 42.4\% in the minimum distance, which validates the superiority of our algorithm. The effectiveness of the experiments in the four scenarios can be seen in Table \ref{four car compare}.

\begin{table}[ht]
\caption{Comparing Four Vehicles at an Intersection}
\label{four car compare}
\begin{center}
\begin{tabular}{ccccccc}
\toprule
& \multicolumn{2}{c}{$f(hz)$} & \multicolumn{2}{c}{$t_{total}(s)$}  & \multicolumn{2}{c}{$d_{min}(m)$}  \\

\cmidrule{2-7}

 & Our & MPC & Our & MPC & Our & MPC \\
\midrule
scenario 1 & 4.93 & 5.00 & 10.86 & 13.49 & 0.43 & 0.74 \\
scenario 2 & 4.05 & 5.00 & 11.78 & 17.12 & 0.41 & 0.78 \\
scenario 3 & 3.95 & 5.00 & 14.25 & 20.59 & 0.43 & 1.20 \\
scenario 4 & 3.59 & 5.00 & 13.42 & 14.54 & 0.38 & 0.52 \\
\bottomrule
\end{tabular}
\end{center}
\end{table}


\section{Conclusion}

This work introduced a multi-vehicle trajectory planning framework based on CE, which balanced traffic efficiency and vehicle safety at intersections. CE recommendations enabled all vehicles to reach a consensus through communication while maintaining the same rationality as an NE, where no vehicle had an incentive to deviate. With V2I, the IM first collected information on each vehicle and then solved an optimization problem to achieve CE among all vehicles. Finally, with the sampled coarse trajectory, a safe corridor was generated by each vehicle, and a dynamically feasible and smooth trajectory was provided. Currently, the algorithm is suitable for scenarios with fixed paths, such as intersections and ramps, and in the future, we will extend it to scenarios with paths changing, such as overtaking and lane merging.





\bibliographystyle{IEEEtran}
\bibliography{IEEEabrv,biblio_traps_dynamics}

\end{document}